\documentclass{article}
\usepackage{amssymb}

\usepackage[preprint]{corl_2025}
\usepackage{times}
\usepackage{tabularx} 
\usepackage[numbers]{natbib}
\usepackage{hyperref}
\usepackage{csquotes}
\usepackage{adjustbox}
\usepackage{booktabs}
\usepackage{multirow}
\usepackage{multicol}
\usepackage{graphicx}
\usepackage{xcolor}
\usepackage{caption} 
\newcommand{\methodname}{DexVLA}
\usepackage{amssymb}
\usepackage{amsmath}
\newcommand{\website}{\url{https://dex-vla.github.io/}}

\title{{DexVLA: Vision-Language Model with Plug-In Diffusion Expert for General Robot Control}\vspace{-0.1in}}

\author{
  \hspace{-1cm}Junjie Wen$^{\ast, 1, 2}$ \quad Yichen Zhu$^{\ast, 1, \dagger}$ \quad Jinming Li$^{1, 3}$ \quad Zhibin Tang$^{1}$ \quad Chaomin Shen$^{2}$ \quad Feifei Feng$^{1}$\\
  \\
  \hspace{-1cm}\large\website
  \vspace{-0.5cm}
}
\begin{document}

\makeatletter
\let\@oldmaketitle\@maketitle%
\renewcommand{\@maketitle}{\@oldmaketitle
    \begin{center}
        \captionsetup{type=figure}
        \centering
        \includegraphics[width=1.0\textwidth]{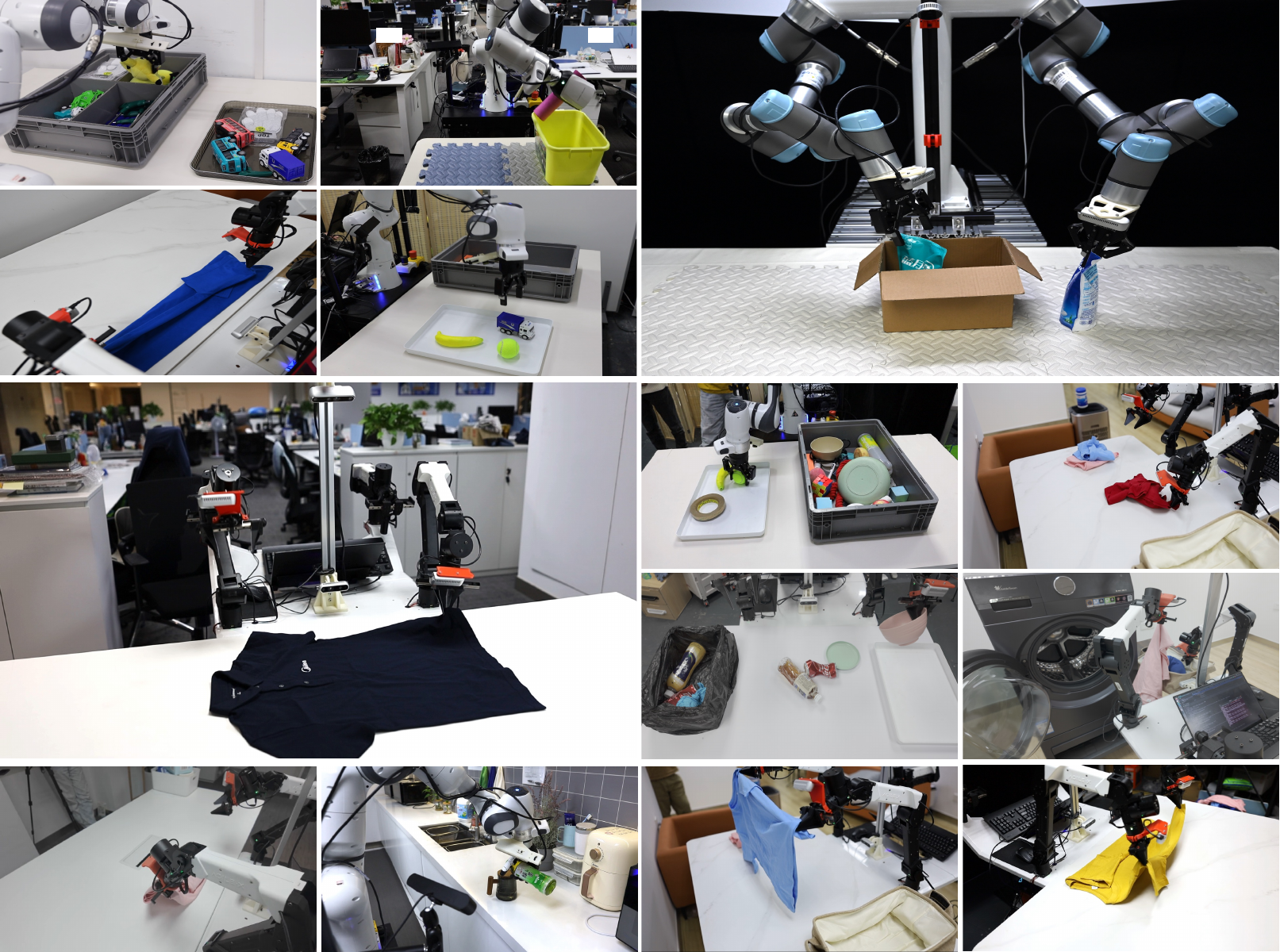}
        \caption{\textbf{Dexterous skills in diverse tasks and scenarios.} Our proposed DexVLA method enables generalized dexterous manipulation across multiple embodiments in diverse scenarios.}\label{fig:framework}
    \end{center}
    \vspace{-1em}
}
\makeatother

\maketitle
\renewcommand*{\thefootnote}{\fnsymbol{footnote}}
\footnotetext{
$^\ast$: denotes equal contribution. $^\dagger$ denotes corresponding author.\\
    $^1$Midea Group, $^2$East China Normal University, $^3$Shanghai University 
}
\renewcommand*{\thefootnote}{\arabic{footnote}}
\begin{abstract}
Enabling robots to perform diverse tasks across varied environments is a central challenge in robot learning. While vision-language-action (VLA) models have shown promise for generalizable robot skills, realizing their full potential requires addressing limitations in action representation and efficient training. Current VLA models often focus on scaling the vision-language model (VLM) component, while the action space representation remains a critical bottleneck. This paper introduces DexVLA, a novel framework designed to enhance the efficiency and generalization capabilities of VLAs for complex, long-horizon tasks across diverse robot embodiments. DexVLA features a novel diffusion-based action expert, scaled to one billion parameters, designed for cross-embodiment learning. A novel embodiment curriculum learning strategy facilitates efficient training: (1) pre-training the diffusion expert on cross-embodiment data, (2) aligning the VLA model to specific embodiments, and (3) post-training for rapid adaptation to new tasks.  We conduct comprehensive experiments across multiple embodiments, including single-arm, bimanual, and dexterous hand, demonstrating DexVLA's adaptability to challenging tasks without task-specific adaptation, its ability to learn dexterous skills on novel embodiments with limited data, and its capacity to complete complex, long-horizon tasks using only direct language prompting, such as laundry folding. In all settings, our method demonstrates superior performance compared to state-of-the-art models like OpenVLA and $\pi_{0}$.


\end{abstract}

\vspace{-0.6cm}
\section{Introduction}
\vspace{-0.3cm}
Enabling robots to perform diverse tasks across varied environments is a central challenge in robotics. Achieving versatility—the ability to solve a variety of tasks across diverse environments, while adapting to language commands, environmental constraints, and unexpected disruptions—is even more demanding. Imitation learning~\cite{lin2024learning,shi2023robocook, lin2024data, zhang2024learning, zeng2022robotic, qin2022dexmv, multimodal_diffusion_transformer}, particularly through vision-language-action (VLA) models~\cite{[pi0, rt-2, kim24openvla, hu2023look, liu2025hybridvla, intelligence2025pi_, kim2025openvlaoft, team2025gemini, bjorck2025gr00t, zhao2025cot-vla, zhao2025vlas, zhen20243d}, has shown promise in enabling generalizable skills.

However, realizing the vision of omnipotent robot foundation models faces persistent challenges. Two key bottlenecks hinder progress: 1) Data scarcity: State-of-the-art models, like OpenVLA~\cite{openvla} and Octo~\cite{octo}, rely on massive datasets like the Open-X Embodiment dataset (4,000 hours)~\cite{o2023open-x} or even larger corpora like the 10,000-hour dataset used by $\pi_{0}$ \& $\pi_{0.5}$~\cite{[pi0,intelligence2025pi_}. Collecting such data through human demonstrations is extremely costly and labor-intensive~\cite{fu2024humanplus, ha2024umi, wu2023tidybot, xiang2020sapien}. (2) Architectural imbalance: current VLA models often prioritize scaling the vision-language model (VLM) component, i.e., OpenVLA uses 7B VLM and $\pi_{0}$ uses 3B VLM. Despite its enhanced visual and linguistic understanding through internet-scale data pretraining, the VLM component remains disconnected from the embodied, sensorimotor context of robotic action.

This paper introduces Plug-in \textbf{D}iffusion \textbf{Ex}pert for \textbf{V}ision-\textbf{L}anguage-\textbf{A}ction models, namely \textbf{DexVLA}, a novel framework designed to enhance the data efficiency and generalization capabilities of VLA for complex, long-horizon tasks across diverse robot embodiments. We achieve this through two key innovations:

\textbf{1) Billion-Parameter Diffusion Expert}: Recognizing the limitations of conventional action experts, particularly in handling cross-embodiment data, we propose a new diffusion-based action expert. The diffusion expert utilizes a multi-head architecture, with each head corresponding to a specific embodiment, enabling effective learning across diverse morphologies.  Furthermore, we scale the model size of the diffusion expert to one billion parameters, a substantial increase from the conventional multi-million parameter scale. This scaling significantly enhances the model's capacity to learn intricate motor skills and control policies from diverse and extensive data.

\textbf{2) Embodied Curriculum Learning}: A three-stage training strategy that progressively learns harder tasks. This is conceptually similar to how human learn, which starts with simple tasks, and then gradually introduces complexity to avoid overwhelming the learner.

\textbf{Stage 1:} The \textit{cross-embodiment pre-training}  stage focuses on learning low-level, embodiment-agnostic motor skills. In this stage, we pre-train only the diffusion expert using cross-embodiment data, without involving the vision-language models.

\textbf{Stage 2:} The \textit{embodiment-specific alignment} is a stage that analogy to ''adapt to your body". Specifically, it bridges abstract vision-language representations to the physical constraints of a specific robot. Remarkably, this stage alone enables the model to complete a variety of tasks, such as shirt folding and bin picking on in-domain objects.

\textbf{Stage 3:} The \textit{task-specific adaptation} aims to make the robot master complex tasks. These tasks include completing long-horizon tasks and generalizing to novel objects.

While the model has learned from diverse robot behaviors and progressively developed dexterous skills for complex tasks, it faces limitations in very long-horizon, contact-rich scenarios such as folding crumpled shirts or executing continuous bin-picking. Prior approaches often rely on high-level policy models; for example, $\pi_{0}$ use SayCan to update instructions every two seconds. In contrast, we propose leveraging the innate reasoning abilities of the VLA model to directly guide robot action. We train the model using demonstrations annotated with \textit{substep reasoning} — for instance, breaking “fold the shirt” into “smooth wrinkles,” “align sleeves,” and “secure folds” — enabling it to learn disentangled action representations that map language sub-instructions to precise motor primitives.

We evaluate DexVLA across diverse embodiments, including single-arm, bimanual, dexterous hand, and mobile bimanual robots, demonstrating its effectiveness on a variety of tasks. DexVLA achieves high success rates on many tasks without task-specific adaptation. For example, it achieves a near full score in folding flattened shirts. It can also learn dexterous skills on novel embodiments with fewer than 100 demonstrations, such as pouring drinks with a dexterous hand and packing on a bimanual robot. Furthermore, when directly prompting VLA model on completing complex, long-horizon tasks like laundry folding, DexVLA outperforms $\pi_{0}$ by a large margin. Importantly, our model is pre-trained on only 100 hours of demonstration data and runs at 60Hz on a single Nvidia A6000 GPU, enabling cost-efficient training and fast inference.




\section{Related Work}
\textbf{Vision-Language-Action models for robot control.} Recent research has focused on developing generalist robot policies trained on increasingly expansive robot learning datasets~\cite{fang2020graspnet, grauman2022ego4d, fu2024humanplus, ha2024umi, ha2023scaling, lin2024data, radosavovic2023real, chi2024universal,geiger2013vision}. Vision-language-action models (VLA)~\cite{openvla,[pi0,pertsch2025fast,wen2024tinyvla,rt-2,zhen20243d,zhang2024grape,guo2025improving,belkhale2024rt-h} represent a promising approach for training such generalist policies. VLAs adapt vision-language models, pre-trained on vast internet-scale image and text data, for robotic control~\cite{yen2020learning}. This approach offers several advantages: leveraging large vision-language model backbones, with billions of parameters, provides the necessary capacity for fitting extensive robot datasets. Furthermore, reusing weights pre-trained on internet-scale data enhances the ability of VLAs to interpret diverse language commands and generalize to novel objects and environments. However, current VLA models do not specifically focus on learning dexterous robotic skills by leveraging the parameters of the underlying VLM. While a few works, such as $\pi_{0}$~\cite{[pi0} and TinyVLA~\cite{wen2024tinyvla}, introduce external action experts to facilitate action learning, their training pipelines still rely on the entire model. Another challenge is that even advanced methods like $\pi_{0}$, despite being capable of completing highly dexterous and long-horizon tasks, require the assistance of a high-level policy, such as SayCan~\cite{ahn2022can}, to decompose tasks into sub-goals. This allows the VLA to complete sub-tasks sequentially. We aim to integrate this high-level planning capability directly into the model itself by training each component of the network with data annotated at the sub-step level. Consequently, our method can complete complex tasks, like laundry folding, without requiring an external high-level policy, making the entire framework more end-to-end and demonstrating significant potential.

\noindent
\textbf{Diffusion models.} Diffusion models~\cite{chen2024yilun,peebles2023scalable,ho2020denoising} have emerged as the dominant approach in visual generation. The Diffusion Policy~\cite{diffusion-policy} successfully applies the diffusion model to robot learning, demonstrating its ability to model multimodal action distributions. Subsequent research has further developed the Diffusion Policy~\cite{aloha_unleashed, wang2024sparse-dp, prasad2024consistencypolicy, multimodal_diffusion_transformer, uehara2024fine, uehara2024feedback, black2023training, black2023zero, dasari2024ingredients, lin2024datascalinglawsimitation, dppo, wang2024inference, liu2022compositional, hu2024video} by applying it to 3D environments~\cite{3d_diffuser_actor, ze20243d, ze2024generalizable, yan2024dnact}, scaling its capabilities~\cite{scaledp}, improving its efficiency~\cite{mail-dp,prasad2024consistencypolicy}, and incorporating architectural innovations. There are a number of works investigating the usage of diffusion VLA~\cite{wen2024tinyvla, [pi0, wen2024diffusionvla}. Although existing models achieve strong performance and generalization on diverse tasks, they predominantly rely on the capabilities of pre-trained vision-language models. This work proposes a paradigm shift towards the diffusion module, demonstrating that a newly designed diffusion-based action expert, coupled with a novel training strategy, enables VLA models to learn from data more efficiently and effectively.
\\
\\
\noindent

\begin{figure*}[t]
    \centering
    \includegraphics[width=0.98\textwidth]{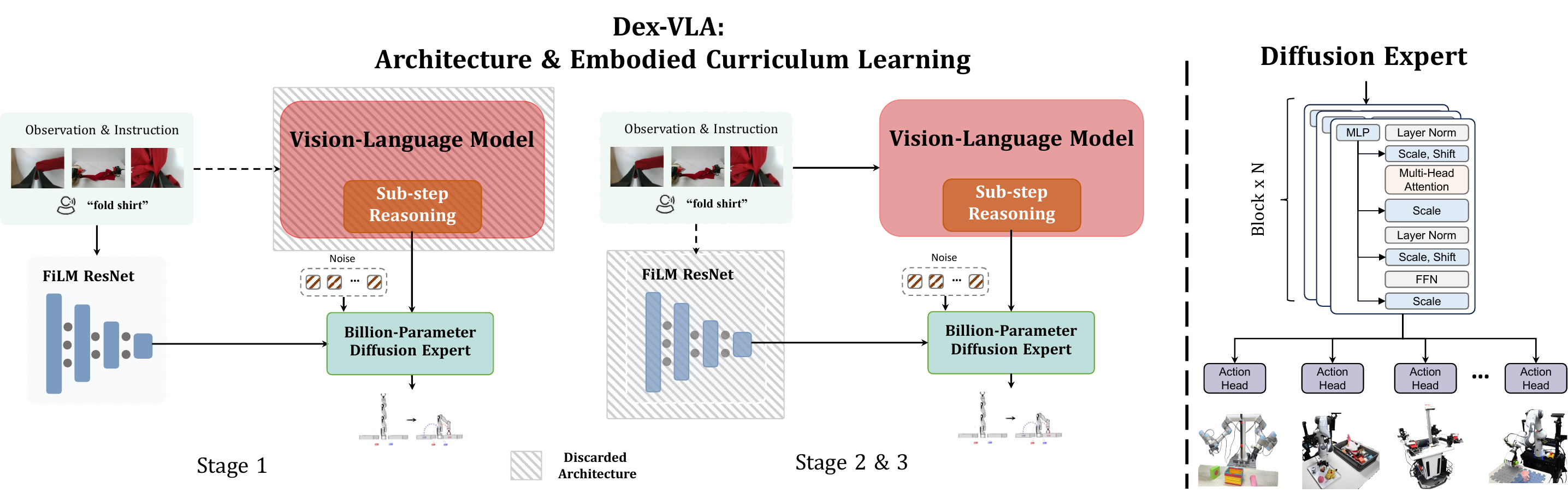}
    \caption{\textbf{DexVLA architecture and embodied curriculum learning.} Our model employs a three-stage training process. \textbf{Stage 1 (left)} trains the Diffusion Expert independently, without the VLM. \textbf{Stages 2 and 3 (middle)} integrate the Diffusion Expert with a VLM, discarding the visual and language components within the expert. \textbf{The Diffusion Expert (right)} uses multiple heads for cross-embodiment learning.}\label{fig:frame_work}
\end{figure*}

\section{Method}
\vspace{-0.3cm}
\subsection{Model Architecture}
\vspace{-0.3cm}
Our~\methodname~model is primarily based on a transformer language model backbone. We use Qwen2-VL~\cite{wang2024qwen2} as the base VLM model. Following the common framework of VLM models, we employ image encoders to project the robot's image observations into the same embedding space as the language tokens. For multiple camera views, these visual tokens are concatenated. The VLM component generates two outputs: reasoning tokens and action tokens. The action tokens are passed through a projection module, consisting of two linear layers with LayerNorm. This module is analogous to the connectors designed in vision-language models like LLaVA~\cite{liu2024visual}, and serves to transform the VLM's embedding space to align with the input requirements of the action expert. The reasoning tokens are injected into the policy model using FiLM layers, which scale and shift the parameters of the projection layers within the policy. Consequently, the model can autonomously generate reasoning and leverage this reasoning within the diffusion expert to guide action generation. The overview is presented in Figure~\ref{fig:frame_work}. 

\textbf{Building diffusion expert.} 
Since action experts dominate the learning process of robot's action, it is essential to design a good neural architecture for better visuomotor policy learning. We utilized the Scale Diffusion Policy (ScaleDP~\cite{scaledp}), a variant of the Diffusion Policy in Transformer architecture, where the largest version of ScaleDP is up to 1B parameters. However, the naive ScaleDP is not designed for cross-embodiment pre-training. We put a multi-head output to enable pre-training on ScaleDP with various robot configurations. Each head is responsible for a single robot configuration. This setup is similar to Octo~\cite{octo}. 

\textbf{Training objectives.} Given a batch of input sequences, the overall training loss is defined as a weighted combination of the diffusion loss ($L_{diff}$) and the next-token prediction loss ($L_{ntp}$), which are $L = L_{diff} + \alpha L_{ntp}$. For all experiments, we set $\alpha = 1$, as we observe that $L_{ntp}$ converges during the early stages of training. Consequently, this setup allows the model to primarily focus on learning robot action prediction based on reasoning and instructions.


\begin{figure*}[t]
    \centering
    \includegraphics[width=\textwidth]{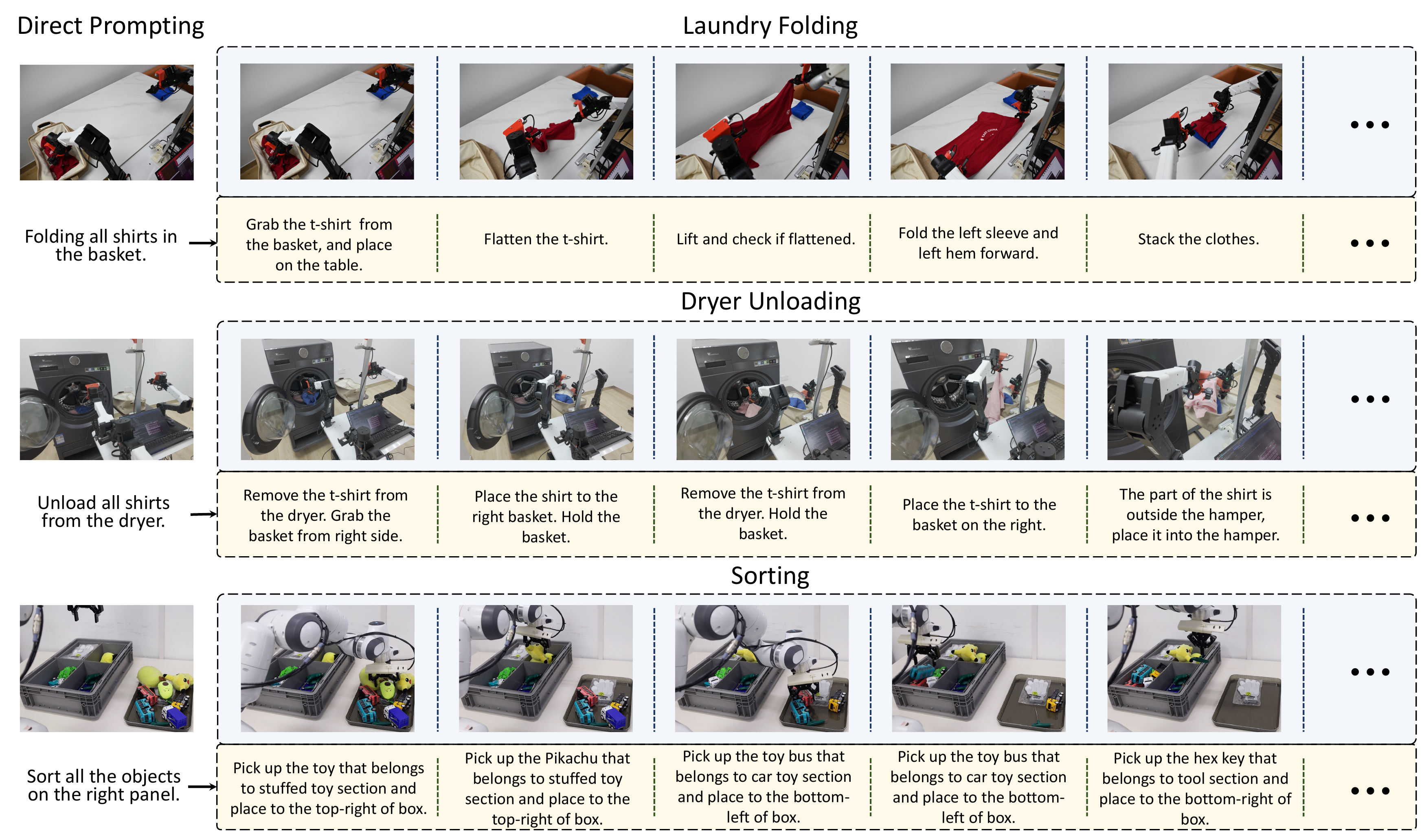}
    \caption{\textbf{Example of direct prompting for long-horizon tasks. } The figure shows three tasks, \textbf{laundry folding (top)}, \textbf{dryer unloading (middle)},  \textbf{sorting (bottom)}. Our DexVLA breaks down raw instructions into sub-steps automatically. Success in these tasks necessitates not only dexterity but also the capacity to decompose direct prompts into implicit multi-step reasoning and to comprehend the visual context.}\label{fig:hard_task_suite}
\end{figure*}

\begin{figure}[t]
    \centering
    \begin{minipage}[t]{0.48\textwidth}
        \adjustbox{valign=t}{%
            \includegraphics[width=\textwidth]{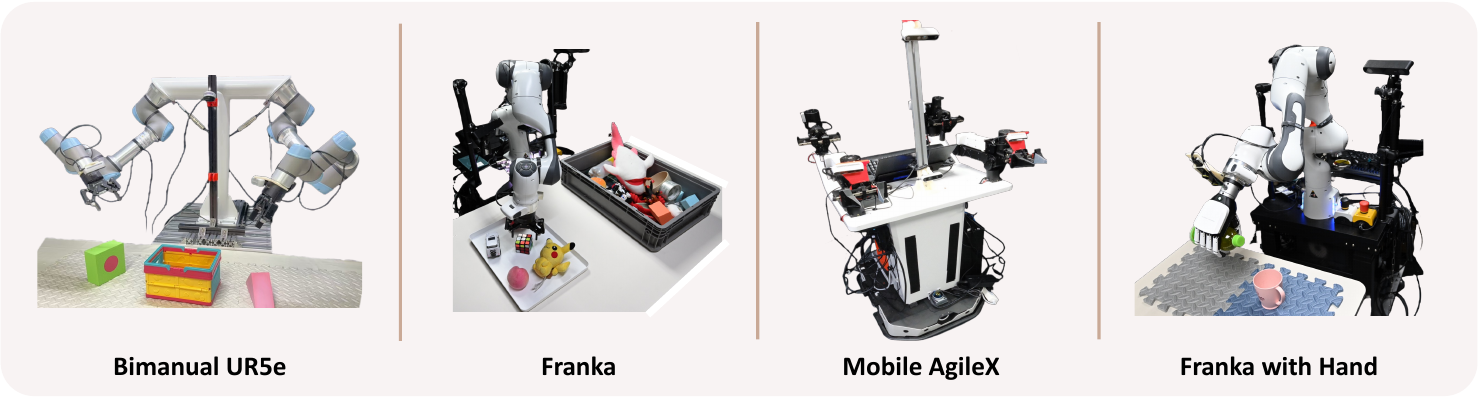}%
        }
        \caption{Our experiment includes various robot types: bimanual UR5e, Franka, bimanual AgileX, and Franka with dexterous hands.}\label{fig:example_robots}
    \end{minipage}\hfill
    \begin{minipage}[t]{0.48\textwidth}
        \adjustbox{valign=t}{%
            \begin{minipage}[t]{\textwidth}
                \centering
                \captionsetup{type=table}
                \caption{\textbf{Detailed hyperparameters for training DexVLA.}}
                \label{tab:hyperparams}
                \resizebox{\textwidth}{!}{
                \begin{tabular}{c|ccc}
                    \toprule
                    Hyperparameters & Stage 1 & Stage 2 & Stage 3 \\
                    \midrule
                    Learning rate & 1e-4 & 2e-5 & 2e-5 \\
                    LR scheduler & Constant & Constant & Cosine \\
                    Weight decay & 0.0 & 0.0 & 0.0 \\
                    Optimizer & \multicolumn{3}{c}{AdamW($\beta_{1}=0.9$,$\beta_{2}$=0.95)} \\
                    Training epochs & 5  & 5 & 5  \\
                    Data & Cross-embodied & Embodied-specific & Task-specific\\
                    \bottomrule
                \end{tabular}}
            \end{minipage}%
        }
    \end{minipage}
\end{figure}

\vspace{-0.3cm}
\subsection{Embodied Curriculum Learning}
\vspace{-0.3cm}
Curriculum learning is an training strategy where a system learns tasks in a progression from simple to complex, mirroring how humans acquire skills. Our three-stage training strategy implements an embodied curriculum, where the policy network first learns generalizable motor skills from cross-embodiment data (Stage 1), then adapts to its specific physical form (Stage 2), and finally refines task-specific behaviors (Stage 3). This mirrors human skill acquisition, where foundational abilities (e.g., grasping) precede specialized expertise (e.g., folding clothes). 

A well-designed training strategy is critical for optimizing deep neural networks. Approaches that align with a network's inherent training dynamics ensure more efficient and effective data utilization. The ~\methodname  targets general robotic control by integrating a VLM with a diffusion expert. Leveraging its modular architecture—which combines two distinct components—we propose a three-stage training strategy that systematically addresses: (1) learning dexterous manipulation skills to enable the model to complete complex tasks; and (2) cross-embodiment learning to adapt the model to diverse robotic platforms.

\textbf{Stage 1: Cross-embodiment pre-training.}
\label{sec:stage1}
The vision-language-action models can be viewed as a composite of two distinct components.  At the top of the architecture lies the vision-language model (VLM), which processes both visual input and language instructions, mapping them into a shared embedding space. This shared space is pre-trained using internet-scale data, enabling a wide range of capabilities, including language understanding, multimodal understanding, and various other vision-text tasks. However, despite its extensive training, the VLM lacks the ability to physically interact with diverse objects in real-world environments.

To effectively pre-train the action expert, we leverage all available data while temporarily decoupling it from the VLM component. This allows us to focus on developing a robust action generation capability independent of language grounding. We use a ResNet-50 as image encoders, aligning with DP~\cite{diffusion-policy} and DistilBERT~\cite{sanh2019distilbert} as a language embedding model. The resulting language embeddings are then integrated into the model using FiLM layers, consistent with previous work~\cite{yell_at_your_robot, brohan2022rt-1}. 

\textbf{Stage 2: Embodied-specific alignment.}
While Stage 1 learns basic motor skills from cross-embodied data, this cross-embodiment learning can potentially compromise performance on the target embodiment, making it unsuitable for real-world deployment. Stage 2 addresses this by training the model with embodiment-specific data, aligning the abstract vision-language representations from the VLM with the diffusion expert. Therefore, we filter the dataset to include only embodiment-specific data, ensuring each sample involves a single embodiment. Mirroring techniques employed in vision-language models like LLaVA~\cite{liu2024visual,liu2024improved}, this stage focuses on aligning the target embodiment's action space with its corresponding camera views and accompanying language instructions. Specifically, we jointly train the VLM model, the projection layer, and the diffusion expert on this embodiment-specific data, while freezing the VLM's visual encoder. This joint training allows the diffusion expert to effectively ground the high-level vision-language understanding from the VLM in the specific motor control space of the target robot. Following Stage 2 training, we observe that the model exhibits proficiency in performing a range of tasks on the target embodiment, such as shirt folding, demonstrating the effectiveness of the embodiment-specific training. 


\begin{figure}[t]
\vspace{-0.4cm}
    \centering
    \begin{minipage}[t]{0.53\textwidth}
        \centering
        \includegraphics[width=\textwidth]{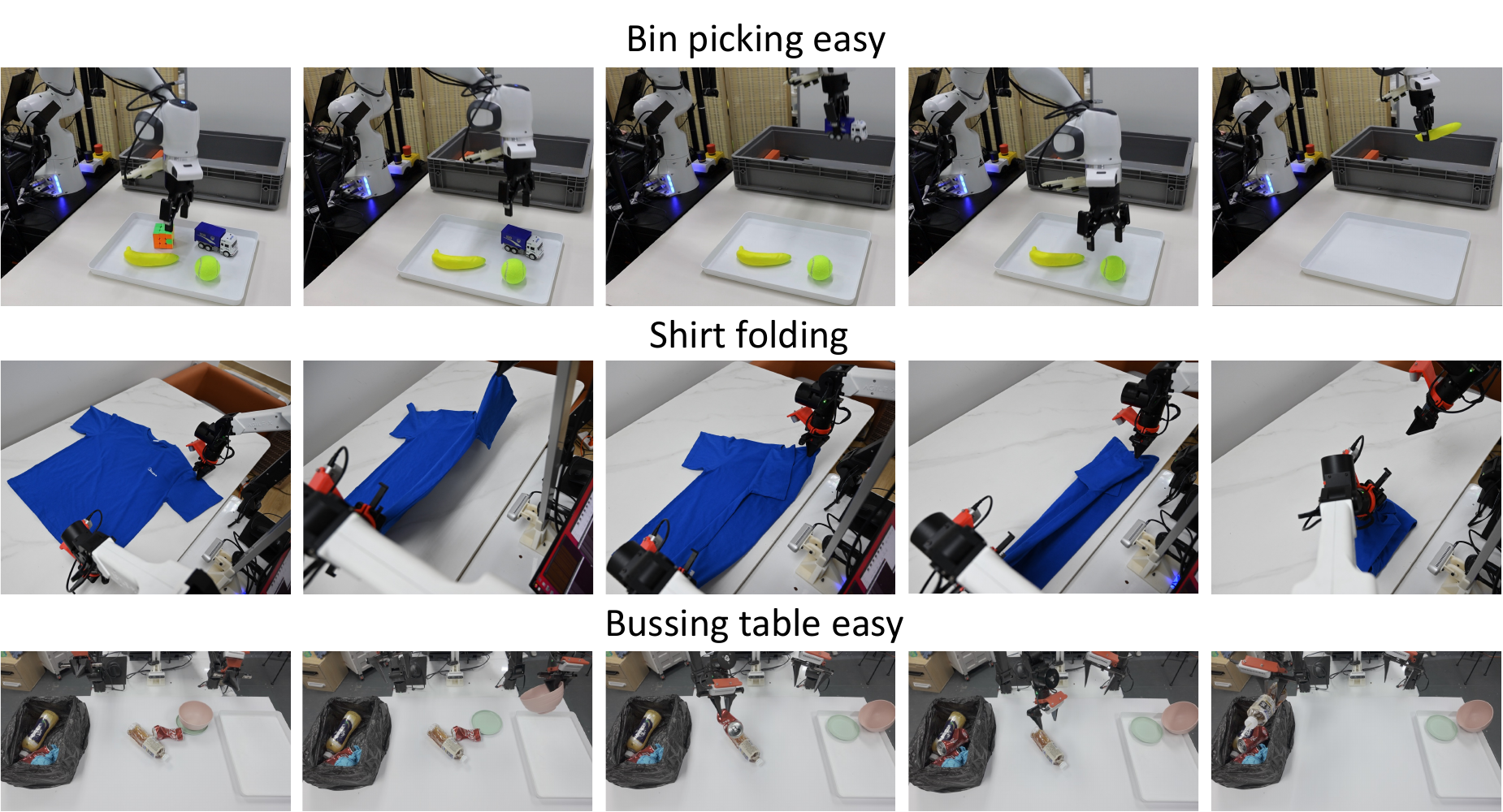}
        \caption{\textbf{Examples of tasks without task-specific adaptation.} We assessed our model's performance after stage 2 training using three tasks: \textbf{bin-picking easy} (top), \textbf{shirt folding} (middle), and \textbf{table bussing easy} (bottom).}\label{fig:easy_task_suite}
    \end{minipage}\hfill
    \begin{minipage}[t]{0.45\textwidth}
        \centering
        \includegraphics[width=\textwidth]{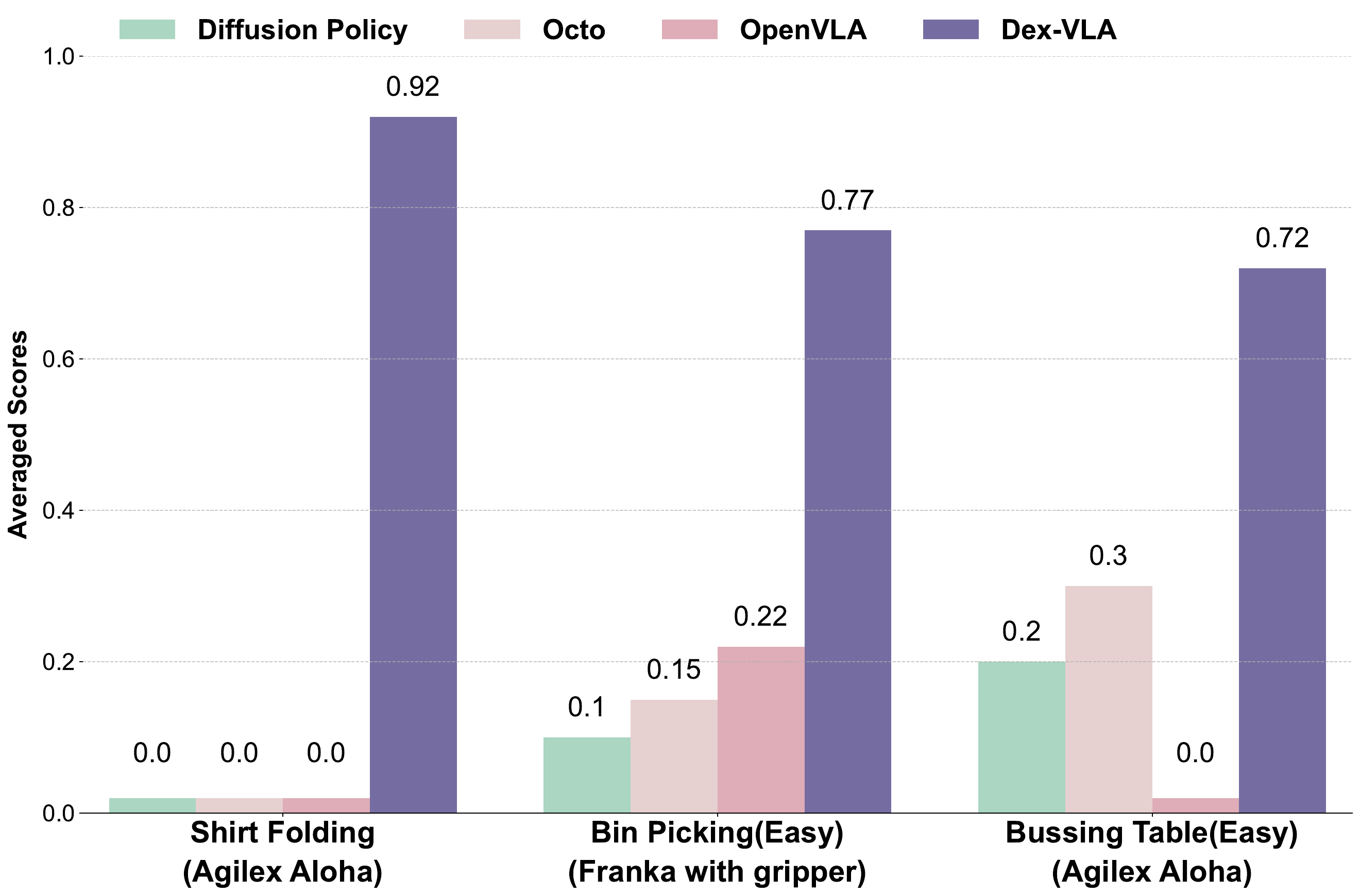}
        \caption{\textbf{Results on tasks without task-specific adaptation.} We compared our model against Octo, OpenVLA, and Diffusion Policy. Performance was evaluated across 10 trials for each model, with scores averaged across these trials.}\label{fig:zero_shot_tasks_results}
    \end{minipage}
\vspace{-0.4cm}
\end{figure}

\textbf{Stage 3: Task-specific adaptation.}
\label{sec:task_specific_adaptation}
This phase refines the model's ability to execute downstream tasks skillfully and fluently, analogous to the post-training stage in large language models where the model is fine-tuned on domain-specific data.  For simpler, less generalization-dependent tasks, such as shirt folding, table bussing, or bin picking with trained objects, task-specific training is unnecessary as the model already performs well.  However, complex, dexterity-demanding tasks require the model to learn fine-grained, context-dependent actions.  Therefore, effective post-training relies on a high-quality dataset of expert demonstrations exhibiting consistent and fluent task execution strategies focused on behaviors that promote successful task completion.
It is worth noting that we utilize sub-step annotated language data in both Stage 2 and Stage 3. However, instead of directly using these sub-step reasoning as instructional input, we employ them as intermediate language output, compelling the model to learn and generate these sub-step language descriptions. This approach has proven highly effective, enabling our model to perform complex, long-horizon tasks such as laundry folding. While other VLAs, like $\pi_0$~\cite{[pi0}, can also complete such tasks, they rely on high-level policy models like SayCan~\cite{ahn2022can} to identify the task state and provide next-step instructions. In contrast, our framework leverages the VLM backbone as an implicit high-level policy model. This allows the model to internally interpret the task's state and inject this understanding into the policy to guide action generation, eliminating the need for an external high-level policy module.

\textbf{Substep reasoning.} A key insight for training the DexVLA is the necessity of decomposing long-horizon tasks (e.g., bussing tables) into sub-tasks. These tasks, often spanning beyond 2 minutes, prove challenging for the diffusion expert to learn effectively from a single language instruction. Therefore, we annotate sub-step instructions within these long-horizon tasks to provide a more structured learning signal. Pre-training with sub-steps is crucial for strong performance. We demonstrate the importance of sub-step reasoning in the Appendix. Our empirical observations show that VLA trained without this pre-training frequently skips critical steps in very long tasks. Sub-step annotations are typically provided every five seconds of the demonstration. We show some examples for substep reasoning in Figure~\ref{fig:hard_task_suite}.

\section{Real World Experiments}



\subsection{Evaluating Model without Task-Specific Adaptation}
\label{sec:no_post_training}
\vspace{-0.3cm}

This section evaluates the model's performance before task-specific adaptation (Stage 3). The evaluated tasks, visualized in Figure~\ref{fig:easy_task_suite}, all use the model with one set of parameters. Detailed task descrptions are listed in Appendix. These tasks vary significantly in trajectory length and complexity, with some requiring high dexterity and intricate manipulation (e.g., shirt folding). We benchmark our approach against OpenVLA~\cite{openvla}, a 7B-parameter VLA model pre-trained on the Open X-Embodiment (OXE)~\cite{o2023open-x} dataset, and Octo~\cite{octo}, a compact 93M parameter model that employs a diffusion-based policy for action generation. We used the open-sourced pre-trained weights for these two models. All baselines are fine-tuned on the same dataset and for the same number of epochs as our Stage 2 training, ensuring a fair comparison. We also compare to the Diffusion Policy~\cite{diffusion-policy}, a strong baseline. Notably, neither Octo nor OpenVLA has previously demonstrated success on tasks of this complexity.



Following $\pi_{0}$, we use a normalized score averaged over 10 episodes per task and method as our evaluation metric.  Detailed scoring rubrics for each task are provided in the Appendix. As shown in Figure~\ref{fig:zero_shot_tasks_results}, DexVLA significantly outperforms all baselines on all tasks without task-specific adaptation.  Notably, baseline methods, including OpenVLA, Octo, and Diffusion Policy, struggled to complete any steps of the shirt folding task, highlighting its complexity.  In contrast, DexVLA achieves a 0.92 point on shirt folding without any task-specific adaptation.  A similar phenomenon is observed in the bin picking and table bussing tasks.  While these challenging tasks sometimes see limited success from the baselines, their overall scores remain low. DexVLA, however, achieves substantially better performance on these tasks.



\begin{figure}[t]
\vspace{-0.7cm}
    \centering
    \begin{minipage}[t]{0.48\textwidth}
    \centering
    \includegraphics[width=1\textwidth]{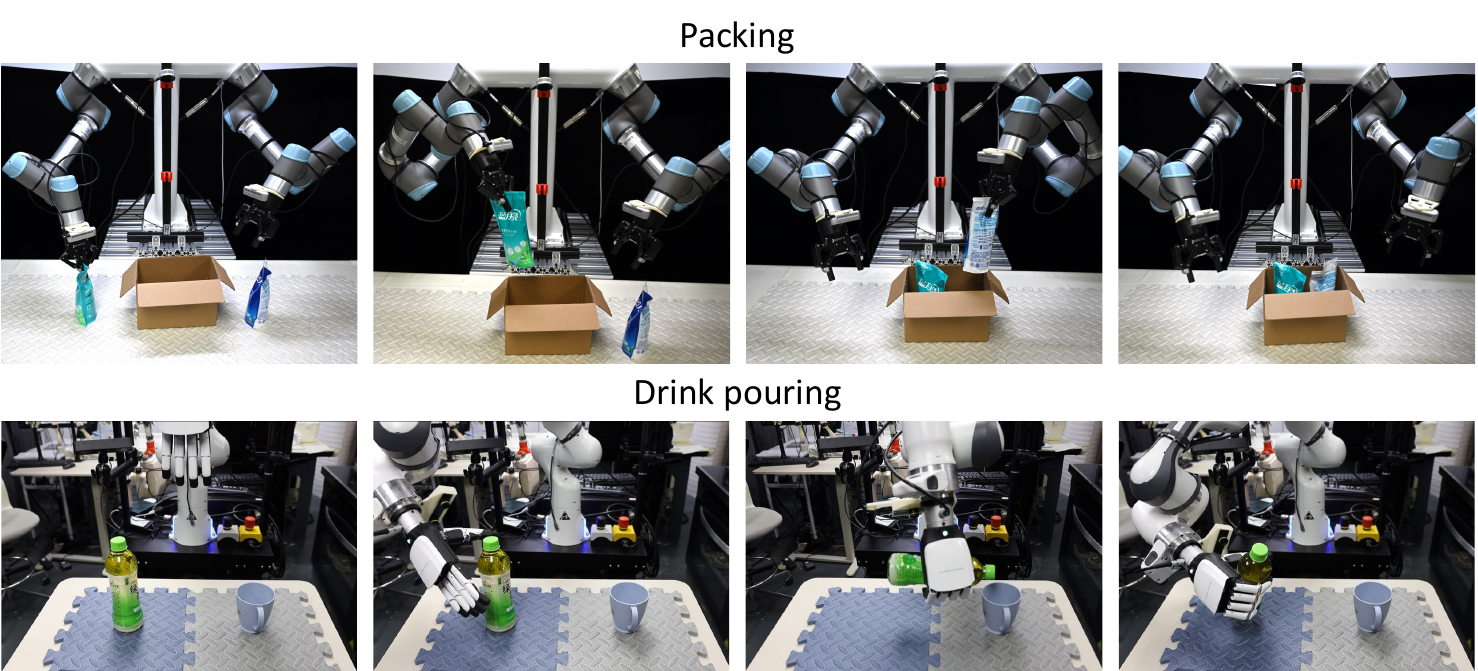}
    \caption{\textbf{Example of tasks for learning dexterous skills on new embodiment.} We evaluate our model on two new embodiments with \textbf{packing (top)} and \textbf{drink pouring (bottom)} tasks, which are not included in stage 1 \& 2 train data.}\label{fig:new_task_suite}
    \end{minipage}\hfill
\begin{minipage}[t]{0.48\textwidth}
    \centering
    \includegraphics[width=0.8\textwidth]{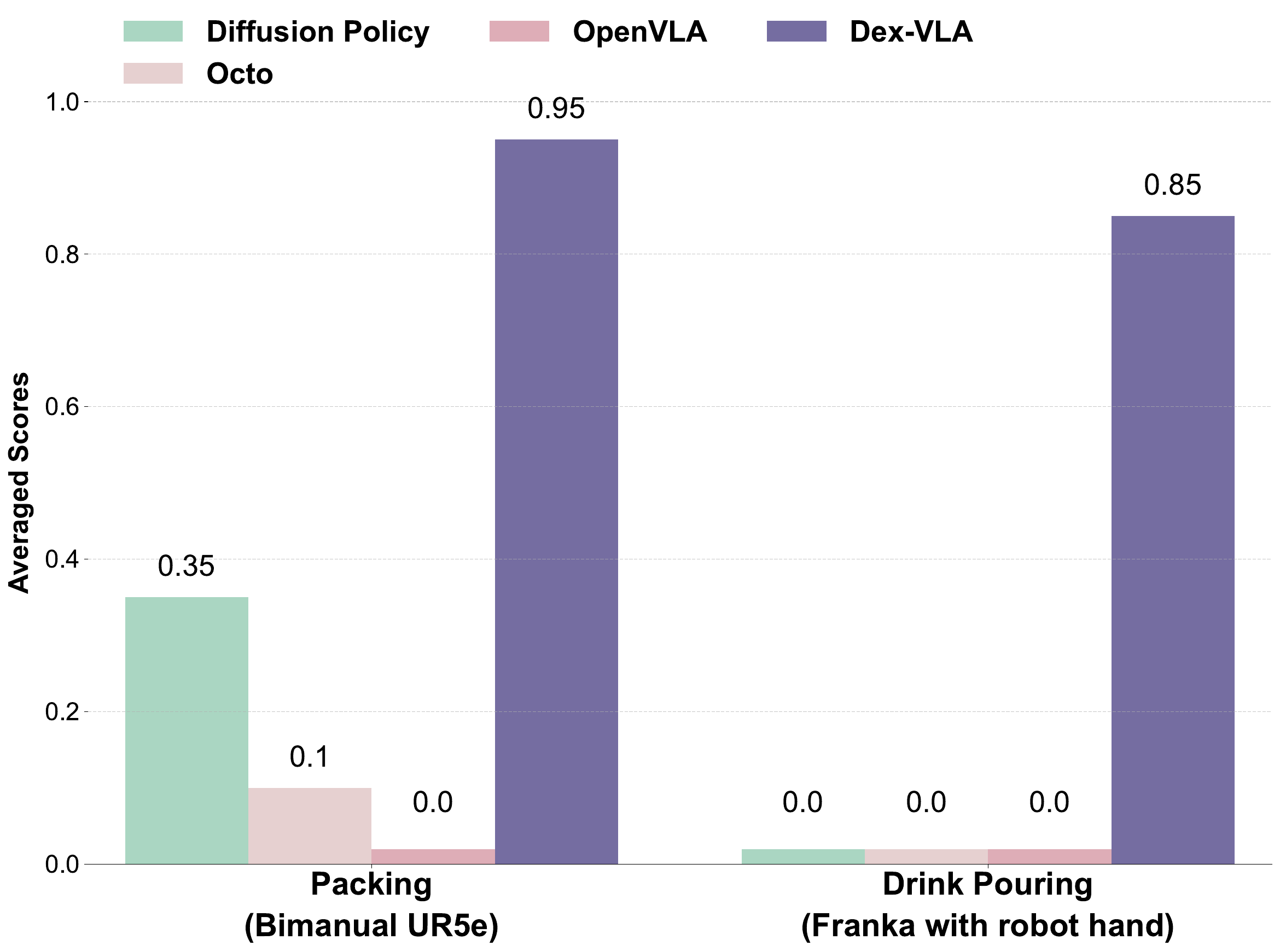}
    \caption{\textbf{Results on learning dexterous skills from new embodiment.} We evaluated our model with four baselines: Diffusion Policy, Octo, and OpenVLA. Diffusion Policy is directly trained on these novel tasks from scratch.}\label{fig:new_tasks_results}
    \end{minipage}
    \vspace{-0.6cm}
\end{figure}

\subsection{Learning Dexterous Skills on New Embodiment}
\label{sec:new_embodiment}
\vspace{-0.3cm}
This section evaluates our model's ability to learn dexterous skills on the new embodiments as shown in Figure~\ref{fig:new_task_suite}. Specifically, the new embodiments are not involved in either Stage 1 or Stage 2 training. We aim to demonstrate the effectiveness of our proposed framework in acquiring new skills quickly on any new embodiment without the necessity of pre-training. Detailed task descrptions are listed in Appendix.

These tasks involve two novel robotic systems absent from training data: 1) \textbf{A Franka arm} integrated with a \textbf{dexterous hand} has 12 DoF, serving as a more complex robotic system than a simple gripper, 2) \textbf{A bimanual UR5e} system featuring humanoid-inspired kinematic design and its articulation fundamentally differs from conventional dual-arm platforms like the AgileX bimanual robot.
We evaluate our approach against the same baselines as in the previous section. This section aims to validate the adaptability of our pre-trained model to new embodiments and tasks. To this end, we directly fine-tune our Stage 2 pre-trained model on the novel tasks. For OpenVLA and Octo, we employ their publicly available checkpoints pre-trained on the OXE dataset. The Diffusion Policy — a method specialized for learning dexterous tasks from limited data — is trained exclusively on the two novel tasks. All methods are trained on individual tasks, and to ensure fairness, each baseline undergoes the same number of training epochs.

Figure~\ref{fig:new_tasks_results} compares the performance of the methods on two novel tasks. For each method and task, we report the averaged scores over 10 trials (detailed scoring criteria are provided in the Appendix). DexVLA achieves an average of 0.90 point across two tasks, while OpenVLA and Octo struggle. DexVLA significantly outperforms Diffusion Policy, achieving a substantial performance lead. These results highlight DexVLA’s ability to efficiently adapt to new embodiments and master complex skills with only 100 demonstrations.  These results are particularly meaningful as our method outperforms both extensively pre-trained VLA models (OpenVLA) and methods specifically designed for learning new tasks (Diffusion Policy).

\begin{figure*}[t]
\vspace{-0.2cm}

  \centering
  \begin{minipage}{0.48\textwidth}
    \centering
    \includegraphics[width=\linewidth]{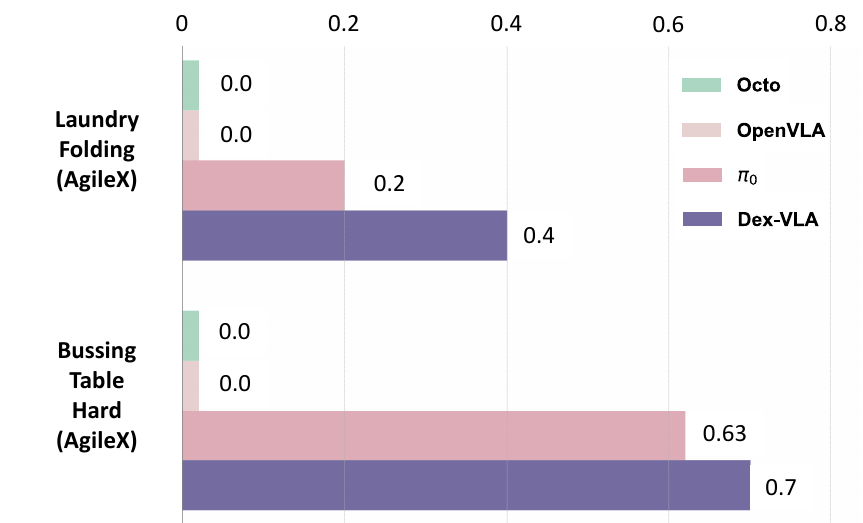}
    \caption{\textbf{Average scores on tasks requiring stage 3 training}. We compared our model against Octo, OpenVLA and $\pi_0$ on laundry folding and bussing table (hard).}
    \label{fig:post_training_results}
    \vspace{-0.6cm}
  \end{minipage}
  \hfill
  \begin{minipage}{0.48\textwidth}
    \centering
    \captionof{table}{\textbf{Ablation study of three stage training}. To evaluate the impact of stage 1 training, we trained DexVLA from scratch without diffusion expert pre-training. Additionally, to assess the effectiveness of stage 3 training, we directly evaluated DexVLA on Laundry Folding without applying stage 3 training.}
    \label{tbl:ablate_scratch}
    \resizebox{0.9\linewidth}{!}{
      \begin{tabular}{ccc|cc}
        \toprule
        Stage 1  & Stage 2 & Stage 3 & Shirt Folding & Laundry Folding \\
        \midrule
        \checkmark & - &-  & 0.0 & 0.0 \\
         - & \checkmark & - & 0.0 & 0.0 \\
        \midrule
        \checkmark & \checkmark & - & 0.92 & 0.0 \\
        \midrule
        \checkmark & \checkmark & \checkmark & 0.92 & 0.4 \\
        \bottomrule
      \end{tabular}
    }

 \centering
 \vspace{0.2cm}
 \caption{\textbf{Ablation results on size of Diffusion Expert.} We reported the average score on the shirt folding task.}
 \vspace{-0.2cm}
 \label{tbl:ablate_head}
 \resizebox{1\linewidth}{!}{
     \begin{tabular}{c|ccc}
       \toprule
       DexVLA & UNet(93M) & Diff. Expert(410M) & Diff. Expert(1B) \\
       \midrule
       Avg. Score & 0.17 & 0.63 & 0.92 \\
       \bottomrule
     \end{tabular}
   }
  \end{minipage}
  
\end{figure*}

\subsection{Complex Long-Horizon Tasks with Direct Prompting}
\label{sec:direct_prompting}
\vspace{-0.3cm}
In this set of experiments, we tackle a range of challenging multi-stage tasks via a combination of task-specific post-training and self-generated reasoning capability. For some of these tasks, data is present in pre-training, but fine-tuning is required to attain mastery. For some, no data is present in pre-training. The tasks in this evaluation, are shown in Figure~\ref{fig:hard_task_suite}, and detailed task descrptions are listed in Appendix. These tasks involve extended-horizon challenges. For instance, laundry folding requires more than 2 minutes to collect a single episode, and the soft, deformable fabric of clothing generates numerous unseen shapes and states, posing significant challenges for recognition and task completion. In the sorting task, the model must pick up 5–8 randomly placed objects in a cluttered scene and relocate them to predefined target positions. We report two tasks, laundry folding and bussing table (hard), in the main text, and discuss the rest of three tasks in the Appendix.

Our evaluation of all models is based on averaged scores over 10 trials, with detailed scoring criteria provided in the Appendix. Conducting comparisons on these tasks is challenging due to the limited availability of prior models capable of operating at this scale. Consequently, we compare our method specifically against OpenVLA, Octo, and $\pi_{0}$. Note that $\pi_{0}$ is pretrained on extensive data and shares the same embodiment as our tested robot. We use direct prompting on all models for fair comparison.

The results, illustrated in Figure~\ref{fig:post_training_results}, demonstrate that DexVLA consistently outperforms all baseline methods. For the most complex task—laundry folding—our method achieves a score of 0.4, showcasing its potential to handle highly complicated scenarios. In comparison, $\pi_{0}$ attains a score of 0.2 when directly prompted with task instructions. On the table bussing task, DexVLA surpasses $\pi_{0}$ by 0.08 points. These findings suggest that DexVLA presents a promising solution for executing complex, long-horizon tasks without relying on an external high-level planner.

\subsection{Ablation Study}
\vspace{-0.3cm}

\textbf{Ablation on three-stage training strategy.} To validate the necessity of this multi-stage training process, we conduct an ablation study in this section. Specifically, we evaluate the DexVLA model under several training conditions: training only Stage 1, only Stage 2, both Stage 1 and Stage 2 combined, and all three stages. Stage 1 consists of pretraining using cross-embodiment data, Stage 2 involves fine-tuning with embodiment-specific data, and Stage 3 enables the model to master more complex tasks. The experimental results are summarized in Table~\ref{tbl:ablate_scratch}. We observe that training solely on Stage 1 or Stage 2 leads to a 0\% success rate in folding tasks. Notably, the absence of Stage 1 training results in the model completely failing to learn any meaningful actions. We hypothesize that the considerable number of parameters in the diffusion expert complicates the optimization process. Thus, Stage 1 serves not only to equip the diffusion expert with foundational action skills but also to "warm up" its parameters, facilitating better comprehension of complex visual cues and language instructions. Additionally, to investigate the contribution of Stage 3 training to performance on more complex tasks, we directly evaluated DexVLA on the laundry-folding task without task-specific adaptation. We observed a significant performance drop from 0.4 to 0. These results highlight that Stage 3 training is essential for the model's success in handling long-horizon and challenging tasks. 

\textbf{Ablation on size of diffusion expert.} Our key contribution is a novel vision-language-action (VLA) model architecture incorporating a diffusion expert, a significantly larger action expert based on a diffusion transformer. However, does the larger diffusion transformer architecture of the diffusion expert (1B) offer advantages over a smaller one?

To address this question, we utilize a tiny UNet-based diffusion policy (93M) and a smaller diffusion expert (410M) as baselines. As shown in Table~\ref{tbl:ablate_head}, the UNet-based action expert performs significantly worse than our method, barely completing the shirt folding task with an average score of 0.17. Empirically, we observed oscillation in the robot's movements with the UNet model compared to our diffusion expert. We hypothesize that the UNet's fewer parameters contribute to interference between different actions in the parameter space, hindering the model's ability to learn the correct actions. The 410M diffusion expert achieved results lower than the 1B model (0.63 versus 0.92), indicating that learning a wide variety of tasks requires a greater number of model parameters.

\section{Conclusion}

This work proposes DexVLA, a novel architecture that leverages vision-language models to learn semantic information and employs a billion-parameter diffusion expert to learn robust and generalizable visuomotor policies. We introduce an embodied curriculum learning strategy, enabling the network to progressively learn from embodiment-agnostic motor skills to complex, embodiment-specific dexterous skills through three training stages.  Furthermore, we incorporate sub-step reasoning, allowing the model to perform very long-horizon tasks without relying on a high-level policy model. Our method is evaluated from multiple perspectives, including its ability to perform complex tasks without task-specific adaptation, fine-tune on new embodiments with limited data, and execute extremely complex, long-horizon tasks without the assistance of a high-level policy model.  



\bibliography{main.bbl}

\clearpage
\clearpage
\newpage
\begin{appendix}

\section{More Experimental Results}
\subsection{Visual generalization.}
Visual generalization is a critical aspect of robot learning.  A well-trained model should not only perform well on in-domain tasks but also generalize to different objects within the same category and to novel scenes. This section presents our visual generalization tests. Specifically, we evaluate shirt folding on a bimanual AgileX and drink pouring using a Franka Emika robot with a dexterous hand. The former task is evaluated without task-specific adaptation, while the latter is trained with 100 demonstrations of the new embodiment. These tasks were also the focus of the experiments presented in Section~\ref{sec:no_post_training} and Section~\ref{sec:new_embodiment}, respectively. For both tasks, we assess visual generalization across two dimensions: novel objects and novel scenes.

\begin{figure}[t]
    \centering
    \includegraphics[width=1\textwidth]{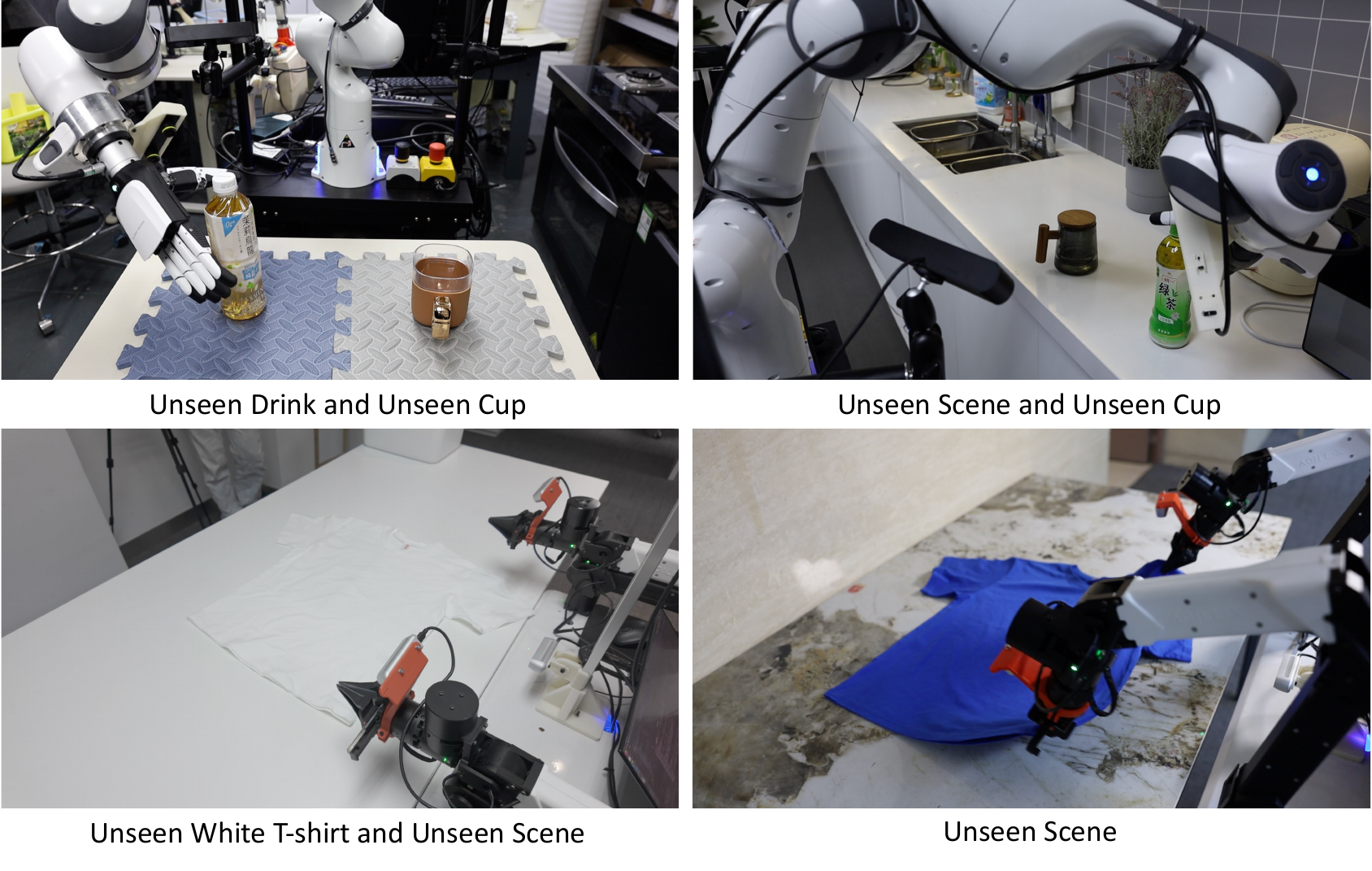}
    \caption{\textbf{Example of visual generalization.} Here lists some visual generalization settings including unseen objects and unseen scenes.}\label{fig:visual_generalization}
\end{figure}

\begin{table*}[t]
  \centering
  \caption{\textbf{Visual Generalizaton for DexVLA}. For each evaluation setting, we report the averaged scores across 3 trials.}
  \label{tbl:visual_generalization}
  \resizebox{1\linewidth}{!}{
      \begin{tabular}{c|c|ccc}
        \toprule
        Task / Generalization & Embodiment &Novel Object & Novel Scene & Novel Object \& Scene \\
        \midrule
         Shirt folding &  Bimanual AgileX & 0.78 & 0.78 & 0.56 \\
         Drink pouring & Dexterous hand& 0.83 & 0.67 & 0.67 \\
        \bottomrule
      \end{tabular}
    }
\end{table*}
For shirt folding, we varied the shirt color (while maintaining size) and altered the background and scene.  For drink pouring, we used unseen cups and bottles, also evaluating the task in different scenes and backgrounds. The results are presented in Table~\ref{tbl:visual_generalization}.  Our experiments demonstrate that DexVLA effectively generalizes to novel visual environments.  As shown in the supplementary video, the model successfully handles even challenging cases, such as folding white shirts on a white table. The examples are shown in Figure~\ref{fig:visual_generalization}.

\subsection{LIBERO simulation experimental results.}
We compare our method on the Libero benchmark. DexVLA uses Stage 1 pre-trained weights. Specifically, the experimental results in Table~\ref{tab:libero} show that DexVLA outperforms all baselines, including $\pi_{0}$ and $\pi_{0}$-FAST, two state-of-the-art VLA methods, demonstrating strong performance on the benchmark. 
\begin{table}[t]
        \centering
        \caption{\textbf{Evaluations results on LIBERO.} We compare our DexVLA with DP and OpenVLA.}
    \resizebox{0.6\linewidth}{!}{\begin{tabular}{l|ccc|c}
    \toprule
    Method & Spatial & Object & Goal & Average \\ 
    \midrule
    DP & 78.3 & 92.5 & 68.3 & 79.7 \\ 
    OpenVLA & 84.7 & 88.4 & 79.2 & 84.1 \\
    $\pi_{0}$-FAST & 96.4 & 96.8 & 88.6& 93.9 \\
    $\pi_{0}$ & 96.8 & 98.8	& \textbf{95.8}	& 97.1 \\
    \textbf{DexVLA} & \textbf{97.2} & \textbf{99.1} & 95.6 & \textbf{97.3}\\
    \bottomrule
    \end{tabular}}
    \label{tab:libero}
\end{table}

\subsection{Training cost of stage 1.}
\begin{table}[t]
  \centering
  \caption{\textbf{Comparison of training cost for train only diffusion expert versus train entire VLA.} Training cost is measured by the number of training epochs completed per hour.}
  \label{tbl:train_cost}
  \resizebox{0.7\linewidth}{!}{
      \begin{tabular}{c|cc}
        \toprule
        Train Method & Train only Diffusion Expert  & Train Entire VLA \\
        \midrule
        Training Cost & 0.89 epoch / hour& 0.32 epoch / hour \\
        \bottomrule
      \end{tabular}
    }
\end{table}
As mentioned in Section~\ref{sec:stage1}, training the entire VLA model from scratch results in failure on nearly all tasks.  Therefore, this section compares the training cost of our Stage 1 (training only the diffusion expert) with that of training the entire VLA model. The test reports the number of training epochs completed per hour. We deliberately keep the same batch size for fair comparison. 

As shown in Table~\ref{tbl:train_cost}, training only the diffusion expert is 2.78 times faster than training the entire VLA model. This is expected, as the VLA model is three times larger than the diffusion expert alone. This highlights that our training strategy is not only effective but also cost-efficient.

\subsection{More discussion on long-horizon tasks with direct prompting.}
In~\ref{sec:direct_prompting}, we previously compared DexVLA with $\pi_0$, Octo, and OpenVLA on laundry folding and bussing table hard. Here, we present additional results against OpenVLA and Octo on more complex tasks. As shown in Figure~\ref{fig:post_training_results}, DexVLA consistently outperforms all baselines, achieving 0.8 points in the dryer unloading task, while Octo and OpenVLA score 0 points. In tasks like sorting and bin picking hard, DexVLA achieves a score nearly 3 times higher than other baselines. Overall, DexVLA demonstrates its versatility in handling complex tasks through our embodied curriculum learning method, scalable diffusion expert, and novel VLA framework. We believe DexVLA presents a promising approach for building VLA systems capable of managing heterogeneous robotics data and mastering intricate manipulation tasks.

\begin{figure}[t]
    \centering
    \includegraphics[width=0.7\textwidth]{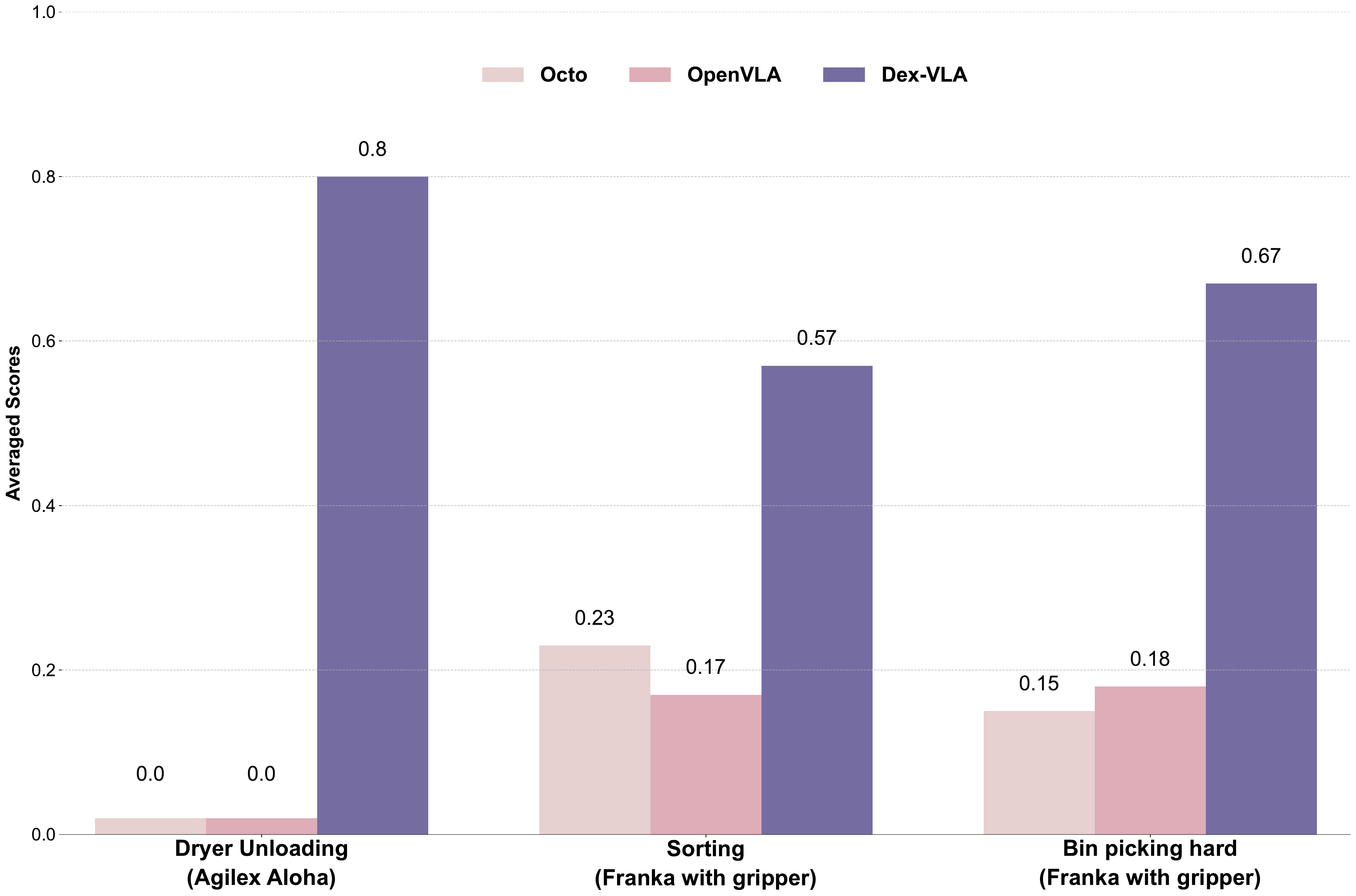}
    \caption{\textbf{Average scores on tasks requiring stage 3 training}. We compared our model against two baselines: Octo and OpenVLA. Averaging scores over 10 trials, our method significantly outperformed both baselines across all tasks. Note that sorting was not included in the pre-training data.}\label{fig:post_training_results}
\end{figure}

\begin{figure}[t]
    \centering
    \includegraphics[width=1\textwidth]{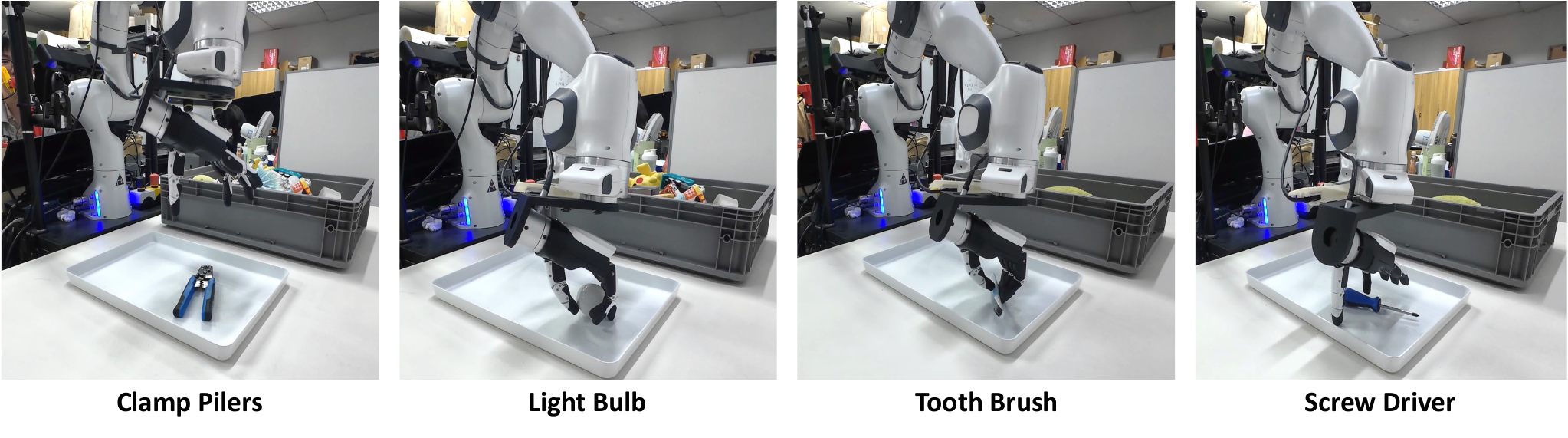}
    \caption{\textbf{Robot setup on zero-shot cross-embodiment transfer from gripper to dexterous hand.} We replace the original gripper with an Inspire dexterous hand and evaluate DexVLA in the same bin-picking environment using unseen objects.}\label{fig:zero_shot_transfer}
\end{figure}

\subsection{Zero-Shot Cross-Embodiment Transfer}
Finally, we pose an intriguing question: can DexVLA perform zero-shot cross-embodiment transfer? To explore this, we take a model trained on a simple two-finger gripper and deploy it, without any further training, on a dexterous five-finger hand. Specifically, we use the Stage 2 pre-trained DexVLA — which was trained on a Franka robot equipped with a Robotiq gripper — and swap in a dexterous hand at test time. Because the gripper has only one degree of freedom, we constrain the dexterous hand to a single degree of freedom as well.

We evaluate on a bin-picking task with 30 novel objects that were unseen during both Stage 1 and Stage 2 training. An illustrative example appears in Figure \ref{fig:zero_shot_transfer}, and a full video demonstration is provided in the supplementary material. Across these 30 objects, we achieve an average success rate of 60\%. While this is slightly below the 67\% success rate attained with the original gripper, it nonetheless underscores the model’s robustness. The performance gap arises from three main challenges: (1) the dexterous hand’s appearance differs markedly from the gripper, forcing the model to generalize its visual feature representations; (2) the wrist camera’s mounting position shifts significantly, requiring adaptation to a new viewpoint; and (3) the difference in hand height changes the effective object grasping height. Although we do not yet control all degrees of freedom needed for fully dexterous manipulation, these results demonstrate that DexVLA’s visual and camera-view representations transfer effectively across embodiments.

\section{Ablation Study}

\subsection{Does training with substep reasoning help?} 
\label{sec:ablation}
\begin{table}[t]
  \centering
  \caption{\textbf{Ablation study of substep reasoning.} The \checkmark in each stage indicates the use of sub-step reasoning data during that stage. We report the average score on the shirt-folding task.}
  \label{tbl:ablate_substep}
  \resizebox{0.4\linewidth}{!}{
      \begin{tabular}{c|ccc}
        \toprule
         Stage 1 & Stage 2 & Averaged Score \\
        \midrule
          & \checkmark & 0.07 \\
         \checkmark &  & 0 \\
        \midrule
         \checkmark & \checkmark & 0.92 \\
        \bottomrule
      \end{tabular}
    }
\end{table}
A key strength of our method is its ability to handle extremely long and complex tasks, such as folding randomly crumpled shirts from a basket.  It also enables the model to complete multi-stage tasks like shirt folding and bin picking without requiring post-training.  Therefore, we now examine the importance of sub-step reasoning. We conducted an ablation study with two setups: 1) The diffusion expert is trained with direct prompting (each task has only one language instruction), while the VLA-diffusion expert is trained with sub-step reasoning. 2) Both stage 1 and stage 2 are trained with direct prompting data.  The results are shown in Table~\ref{tbl:ablate_substep}. Training the diffusion expert with direct prompting, even for a relatively simple task like shirt folding, reduces the averaged score from 0.92 to 0.07.  Furthermore, removing sub-step reasoning from both stages results in a complete failure (0 score). This is a significant observation. It suggests that learning long-horizon tasks within a shared parameter space can sometimes lead to conflicts. We hypothesize that sub-step reasoning allows the model to learn a more disentangled action space, similar to mapping a continuous action space to a discrete one~\cite{wu2024discrete}. This effectively segments the shared parameter space, allocating a smaller set of parameters to each substep~\cite{wang2023fleet, wang2024poco, wang2024scaling}. This avoids parameter conflicts, leading to improved performance and generalization.

\subsection{Explicit versus implicit sub-step reasoning.}

\begin{table}[t]
\centering
\caption{\textbf{SayCan versus Substep Reasoning for DexVLA.} We replace DexVLA's substep reasoning with SayCan and evaluate how this change affects overall performance.}
\label{tbl:saycan_compare}
\begin{tabularx}{0.6\linewidth}{c|cc}
    \toprule
    Tasks/Models & SayCan & Substep Reasoning \\
    \midrule
    Bussing Table (Hard) & 0.58 & 0.70 \\
    \bottomrule
\end{tabularx}
\end{table}
The $\pi_0$ utilizes the additional SayCan to explicitly plan substeps, while DexVLA generates substep reasoning implicitly. Here, we evaluate the impact of these two reasoning approaches on overall performance. Specifically, we replace DexVLA’s implicit substep reasoning with SayCan's. As shown in Table~\ref{tbl:saycan_compare}, models using implicit substep reasoning significantly outperform those relying on SayCan. The key advantage of implicit substep reasoning is its ability to disentangle the action space across different features, allowing DexVLA to learn more effectively. Additionally, SayCan updates instructions at a fixed frequency (every two seconds), which can result in redundant or repeated states in certain scenarios. In contrast, DexVLA’s substep reasoning adaptively segments the state space over the course of long-horizon tasks, contributing to its superior performance.

\section{Task Suite and Evaluation Protocol}

\subsection{Task suite.}
\label{sec:x_tasks}
These tasks are evaluated in Section~\ref{sec:no_post_training}.
\begin{itemize}
    \item \textbf{Shirt folding (Bimanual AgileX)}: The shirt is placed flattened on the table, and the robot is asked to fold a t-shirt. We evaluate two shirts, a yellow shirt of medium size and a blue shirt of large size. 
    \item \textbf{Bin picking easy (Franka with gripper)}: The model needs to pick up all items from the right panel to the left tray. All items are seen in the dataset. 
    \item \textbf{Bussing table easy (Bimanual AgileX)}: The robot must clean a table, place dishes and cutlery in a bin, and trash into a trash bin.
\end{itemize}

These tasks are evaluated in Section~\ref{sec:new_embodiment}.
\begin{itemize}
    \item \textbf{Drink pouring (Franka with dexterous hand)}: The drink is placed on the right of the table and a cup is placed on the left. The robot needs to grab the drink and pour it into the cup. This task includes 100 demonstrations. 
    \item \textbf{Packing (Bimanual UR5)}: The robot is asked to pick up objects on both sides and place them into the box for packing. This task includes 100 demonstrations. 
\end{itemize}

These tasks are evaluated in Section~\ref{sec:direct_prompting}.
\begin{itemize}
    \item \textbf{Laundry folding (Bimanual AgileX)}: This task requires a static (non-mobile) bimanual system to fold articles of clothing. The clothing items start in a randomized crumpled state in a bin, and the goal is to take out the item, fold it, and place it on top of a stack of previously folded items. The randomized initial configuration of the crumpled laundry presents a major challenge since the policy needs to generalize to any configuration. This task is present in pre-training.
    \item \textbf{Dryer unloading (Bimanual AgileX)}: Here, the AgileX mobile robot has to take the laundry out of a dryer and place it into a hamper. This task is present in pre-training.
    \item \textbf{Sorting (Franka with gripper)}: The model needs to pick up all items from the right panel, and place them in the correct subsection of the left tray. The left tray is divided into four subsections. All new objects belong to the same category as these four sections. This task includes 200 demonstrations. This task is not in the pre-training data. 
    \item \textbf{Bin picking hard (Franka with gripper)}: The model needs to pick up all items from the right panel to the left tray. Unlike the easy version, all objects are new and only present at test time. This task is present in pre-training.
    \item \textbf{Bussing table hard (Bimanual AgileX)}: The robot must clean a table, place dishes and cutlery in a bin, and trash into a trash bin. Unlike the easy version, all objects are new. In particular, we use dishes with unseen colors and trash with different appearances. This task is present in pre-training.
\end{itemize}



\subsection{Evaluation protocol.}
Each task is evaluated across 10 trials and reported averaged scores. For each task, we list the detailed scoring criterion as follows.
\begin{itemize}
    \item \textbf{Lanudary folding (Bimanual AgileX)}: This task is scored out of 4 and we evaluate 5 shirts in total including 2 middle size and 3 small size. We perform two trials for each item, and the items left to be evaluated starting randomly crumpled in a laundry bin (while previously evaluated items start in a fold). One point is given for picking an item out of the bin and putting it on the table. Another point is given for flattening the shirt or shorts. A third point is granted for folding the shirt or shorts. A final point is given for either placing the item in the corner of the table (if it is the first item evaluated), or stacking it onto an existing stack of folded clothes. This evaluation metric is followed $\pi_{0}$. 
    \item \textbf{Shirt folding (Bimanual AgileX)}: This task is scored out of 3. We perform two trials for each item, and the items are flattened on the table. One point is given for double vertically fold. Another point is granted for a double horizontal fold. A final point is given for pushing the folded shirt to the right blank area.
    \item \textbf{Bussing table (Bimanual AgileX)}: This task is scored out of 3-4 where there are 3-4 objects on the table in both \textbf{easy and hard version}. The main difference is the objects that appeared in the hard version are unseen. A point is given for each correctly sorted object.
    \item \textbf{Dryer unloading (Bimanual AgileX)}: This task is scored out of 2 where there are 2 crumpled shirts in the dryer. A point is given for pick up a shirt and place into the hamper.
    \item \textbf{Sorting (Franka with gripper)}: This task is scored out of 5-8 where there are 5-8 objects on the table. There are four kinds of objects in total, a point is given for each correctly sorted object.
    \item \textbf{Drink pouring (Franka with dexterous hand)}: This task is scored out of 2. A point is given for grab the bottle and pour to the cup. Another point is granted for place down the bottle.
    \item \textbf{Bining picking (Franka with gripper)}: This task is scored out of 4-5 where there are 4-5 objects on the table. The main difference is the objects that appeared in the hard version are unseen. A point is given for each correctly picked and placed object.
    \item \textbf{Packing (Bimanual UR5e)}: This task is scored out of 2 where there are 2 objects on the table. A point is given for each correctly picked and placed object.
\end{itemize}

\begin{figure}[t]
    \centering
    \includegraphics[width=0.50\textwidth, trim=0pt 0pt 0pt 0pt, clip]{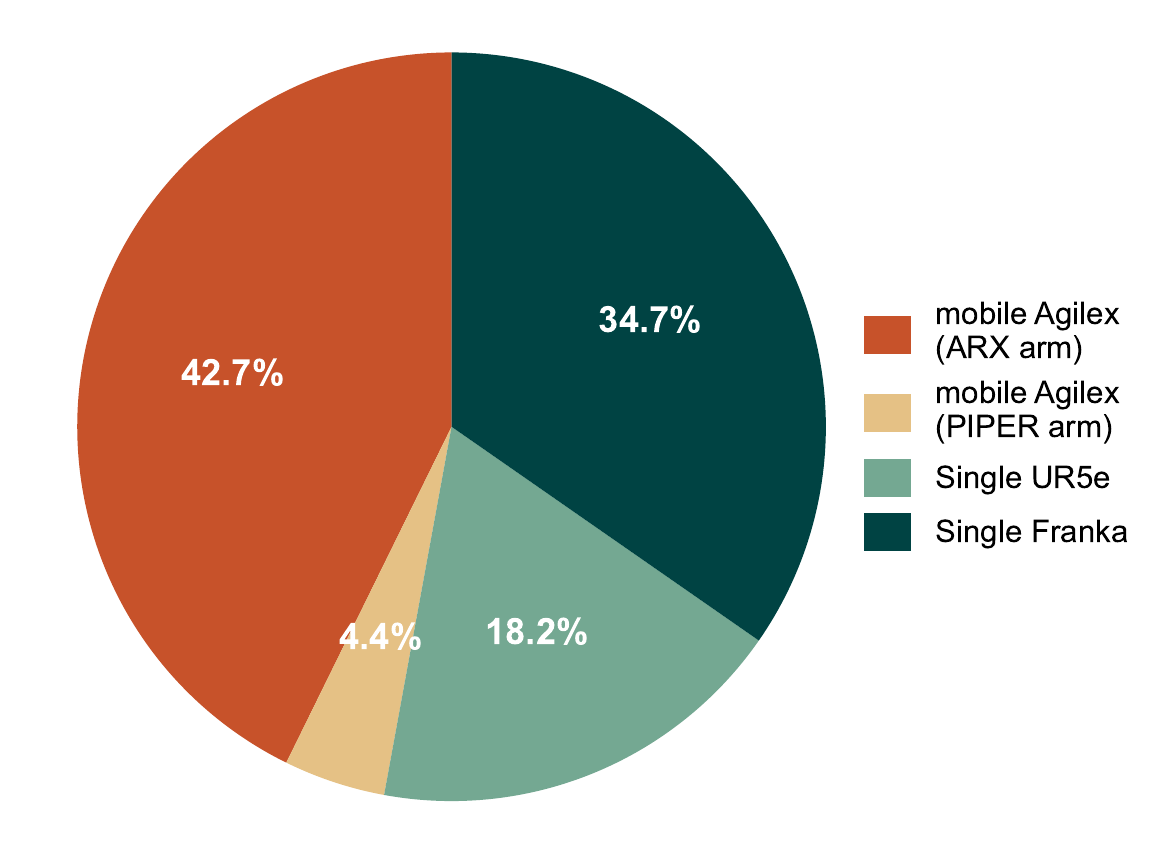}
    \caption{Overview of our dataset for stage 1 training.}\label{fig:data_allocation}
\end{figure}

\begin{figure}[t]
  \centering
  \includegraphics[width=1\linewidth]{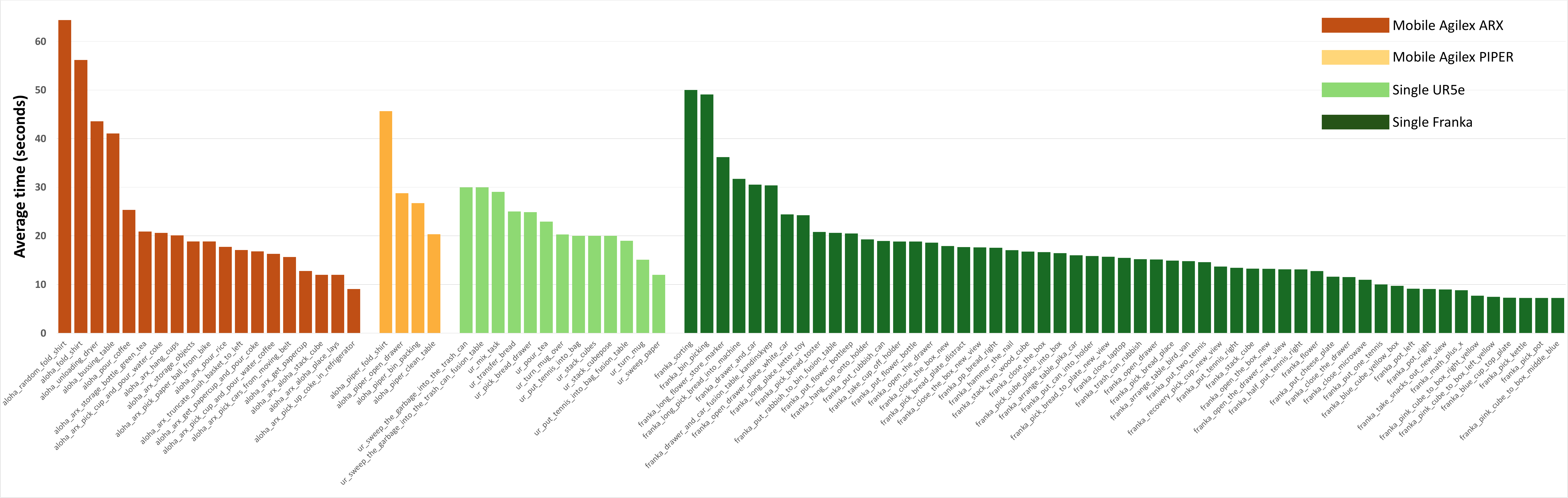}
  \caption{Average task length(seconds) of 91 tasks. \label{fig:91_tasks}}
\end{figure}

\section{More Implementation Details}

\subsection{Robot setup.} 
Our evaluation is conducted on four different robot configurations across 10 tasks. These setups are summarized in the following list and visualized in figure~\ref{fig:example_robots}.

\begin{itemize} 
\item \textbf{Franka with gripper.} A Franka Emika robot with 7 degrees of freedom, equipped with a Robotiq parallel jaw gripper. Data is collected at 15Hz. 
\item \textbf{Franka with dexterous hand.} An Inspired Dexterous Hand mounted on a Franka Emika robot. The camera setup is identical to the gripper version, with a total of 12-dimensional configuration. Specifically, SE(3) end-effector pose (3D position + 3D orientation)
plus 6-dimensional hand joint space. Data is collected at 15Hz. 
\item \textbf{Bimanual UR5e.} Two UR5e robots, each with a Robotiq parallel jaw gripper and a wrist-mounted camera. A top-down camera is positioned between the two arms. This setup has a total of three camera views and a 14-dimensional configuration and action space. Data is collected at 10Hz. 
\item \textbf{Bimanual AgileX.} Two 6-DoF AgileX arms, each with a wrist-mounted camera and a base camera. This setup has a 14-dimensional configuration and action space, supported by three cameras in total. Data is collected at 10Hz. 
\end{itemize}

Our setup includes two Franka Emika robots: one equipped with a gripper and the other with a robot hand.  Both Franka robots utilize the same camera configuration, consisting of a ZED 2 camera positioned on both the left and right sides, as well as a ZED Mini wrist camera mounted on the robot itself.  Our bimanual UR5e robot uses a single top-mounted Intel RealSense L515 camera and two Intel RealSense 435i cameras attached to the wrists.  Finally, our mobile AgileX platform has two Intel RealSense 435i wrist cameras and a top-mounted Intel RealSense 457 camera.  Although the mobile AgileX image includes a front camera, it was not used during either training or inference. 

\subsection{Sub-step reasoning and data acquisition.}
Training with sub-step reasoning is crucial for Dex-VLA to complete long-horizon tasks without a high-level policy model. We present an ablation study on the importance of sub-step reasoning in Section~\ref{sec:ablation}. Acquiring this data presents two key challenges: obtaining language instructions and segmenting videos with corresponding annotations. We address these challenges with the following strategy.

For object-level tasks (e.g., bin picking, sorting, table bussing), object identification is key. We leverage Grounding-Dino and DINOv2 to annotate object bounding boxes and names, along with the gripper's bounding box.  We then calculate the intersection over union between the gripper and object bounding boxes to determine grasp success. For long-horizon single-object tasks (e.g., fold one shirt), the challenge lies in task segmentation. We created a comprehensive list of potential sub-step reasoning, focusing on major steps lasting at least five seconds each to avoid excessive sub-division. We then used Google Gemini 2.0, providing it with the sub-step list, to segment the videos and select the corresponding reasoning from the list. This proved effective and efficient for labeling.  We only manually checked the Stage 3 training data, as this stage requires higher-quality annotations. This annotation strategy makes our approach feasible.

\subsection{Architectural details.} In this section, we provide a full description of the model architecture. Dex-VLA can be split into two parts, VLM backbone originates from Qwen2-VL~\cite{wang2024qwen2} and diffusion expert. We use Qwen2-VL 2B which is powerful and efficient. Regarding our Diffusion Expert, the total number of parameters for this model is 1 billion parameters. We use 32 layers, with the hidden stage of 1280, and a number of heads of 16. During Stage 1, we only pre-train the diffusion expert with random initialized ResNet-50 to process images and off-the-shelf Distilbert~\cite{sanh2019distilbert} to encode language instructions. Because the original diffusion policy model does not support cross-embodiment training, we adopted a multi-head structure similar to Octo~\cite{octo}. Each embodiment is assigned a unique MLP head. The diffusion expert is trained using the similar settings of our Dex-VLA. In particular, we use the image resolution of 320 $\times$ 240, with three camera views. Each image is processed independently to a ResNet-50. We use the strategy as in RT-1~\cite{brohan2022rt-1} to initialize the FiLM layers.

\subsection{Training data details.}
As shown in Figure~\ref{fig:data_allocation}. Our dataset comprises approximately 100 hours of collected data spanning 91 distinct tasks. The majority of this data was collected using two robot platforms: the Agilex (ARX arm) (42.7\%) and the single Franka Emika robot (34.7\%). The ``ARX arm'' and ``PIPER arm'' represent two distinct robotic arm configurations, both featuring six degrees of freedom (6-DoF) but differing in their kinematic structures and operational characteristics.

Figure~\ref{fig:91_tasks} provides a summary of all 91 pretraining tasks across four embodiments. The Y-axis represents the average duration (in seconds) recorded for each task in the training data, while the X-axis lists all 91 tasks. It is clear that most tasks in the dataset are short-horizon, whereas the evaluated tasks are long-horizon, highlighting a notable distributional difference between the pretraining data and the evaluated tasks.

\end{appendix}
\end{document}